# Robot Manipulator Control with Inverse Kinematics PD-Pseudoinverse Jacobian and Forward Kinematics Denavit Hartenberg

Indra Agustian [a, *], Novalio Daratha [a], Ruvita Faurina [a], Agus Suandi [a], Sulistyaningsih [b]

[a] *Faculty of Engineering*
*University of Bengkulu*
*Jl. W.R Supratman Kandang Limun, Muara Bangkahulu*
*Bengkulu, Indonesia*
[b] *Pusat Penelitian Elektronika dan Telekomunikasi*
*Lembaga Ilmu Pengetahuan Indonesia*
*Jl. Sangkuriang, Komplek LIPI Gd. 20, Dago, Kecamatan Coblong*
*Kota Bandung, Jawa Barat, Indonesia*

**Abstract**

This paper presents the development of vision-based robotic arm manipulator control by applying Proportional Derivative-Pseudoinverse Jacobian (PD-PIJ) inverse kinematics and Denavit-Hartenberg forward kinematics. The task of sorting objects based on color is carried out to observe error propagation in the implementation of manipulator on real system. The objects image captured by the digital camera were processed based on HSV-color model and the centroid coordinate of each object detected were calculated. These coordinates are end effector position target to pick each object and were placed to the right position based on its color. Based on the end effector position target, PD-PIJ inverse kinematics method was used to determine the right angle of each joint of manipulator links. The angles found by PD-PIJ is the input of DH forward kinematics. The process was repeated until the end effector reached the target. The experiment of model and implementation to actual manipulator were analyzed using Probability Density Function (PDF) and Weibull Probability Distribution. The result shows that the manipulator navigation system had a good performance. The real implementation of color sorting task on manipulator shows the probability of success rate cm is 94.46% for euclidean distance error less than 1.2 cm.

## I. INTRODUCTION

Robot manipulator is one of the most popular robot technologies used in industry today. Manipulator robots can be designed and programmed as needed to perform repetitive tasks with speed, accuracy, precision, and endurance that far exceed human capabilities. Like the human hand, the robot arm manipulator moves the end effector from one place to another, picks up and places objects, welds, writes, paints, and so on.

Robot manipulator or usually just called manipulator, is a mechanical arm consisting of frames (links) that are interconnected via hinges or joints that allow relative movement between two successive frames [1], [2]. The main problem of the manipulator is how the mathematical relationship between the base position, the angle of the joint, and the position of the end effector, either in a stationary or dynamic state. The second problem is the design and control of manipulator systems with specific mechanical structures, specific to implementation or specific purposes.

The first problem can be solved with kinematics, which is a branch of mechanics that studies the motion of an object or system without considering the mass and force given [3]. The kinematics problem itself consists of two main parts, namely forward kinematics and inverse kinematics. In forward kinematics, the end effector position is a function of the joining angles, simple and has only one solution. Whereas the inverse kinematics, the joint angles are a function of the position of the end effector position [4]. An inverse kinematics solution can be more than one, called redundant. It can also have no solution called, called a singularity. Therefore, the inverse kinematics problem becomes more complex and complicated and is proportional to the number of degrees of freedom/DOF (Degree of Freedom).

The solution to the second problem depends on the choice of mechanical structure used and the target implementation. There are many manipulator modes based on mechanical structures such as Cartesian Robot, SCARA (Selective Compliance Articulated Robot Arm), Articulated Robot, Parallel Robot, Cylindrical Robot [5], [6]. There is almost no problem for forward kinematics, but for inverse kinematics, each mechanical structure requires a different approach to get the maximum performance.

One of the manipulator implementations studied by several researchers is the manipulator for sorting objects based on certain characteristics. Compared to human operators, sorting activities can be more effective when using manipulators. Since these activities are repetitive, manipulators with the right design can have better accuracy, precision, and durability.

In this research, the manipulator navigation system and its implementation are designed and discussed to carry out sorting activities based on the color of the digital image input. Research [7] - [10] designed a robot arm using a color sensor as the manipulator's motion control input, but the color sensor did not have the ability to obtain object location data. While research [8] - [13] used image input from a digital camera to obtain color data and object location at once.

Djajadi [11] focused on object detection system using Hough Transform method and sorting implementation using Mitsubishi Movemaster RV-M1 Robot Manipulator. Djajadi did not mention the inverse kinematic method that was applied, the starting and ending location of the object was fixed and had been determined, the joint angle had also been calculated manually. Szabó [12] designed a color-based sorting application and used the Lynxmotion AL5A manipulator with the same concept as Djajadi. Ata (2013) [13] developed a color-based sorting system with a 5-DOF manipulator based on digital image input, the inverse kinematic was solved by analytical methods. However, this analytical method approach becomes difficult and slow if it is developed for more complex kinematics problems [14]. Tsai (2015) [15] proposed a hybrid-switched reactive visual servo navigation system to the 5-DOF manipulator for sorting by color and shape, also the image-based visual servoing (IBVS) method for the effector function of retrieving objects and position-based visual servoing (PBVS) for initial-position navigation to targets. In contrast to studies [8] - [10] which used a single camera, Tsai used two cameras. The first camera is for PBVS and the second camera for IBVS. Jia [16] used the Phantom X Reactor manipulator, one camera, and the Pseudoinverse Jacobian method for inverse kinematics. In his article, Jia does not explain the inverse kinematics mechanism used, but rather focuses on detection systems and electronic hardware. Abad [17] proposed a Fuzzy Logic Based Joint (FLJC) control system on a 6-DOF manipulator for color-based sorting applications with one camera. The main difficulty of FJLC is determining the membership function for each state and the optimal fuzzy rules for each angle of the manipulator.

In this study, a software and hardware manipulator package were designed to automatically sort objects based on color. A manipulator that was designed in a serial model with a mechanical articulated robot structure [5] adopted the MeArm Robot mode [18] and a microcontroller as a servo control interface. The manipulator motion control is based on visual information (digital image) from a single USB camera as the basis for the movement of the end-effector manipulator. The image from the camera is extracted using the HSV color segmentation approach [19], [20] to get color and location information for the target object. The colors detected are only red, green, and blue. Forward kinematics uses the Denavit-Hartengberg (DH) method [21] for inverse kinematics with the Jacobian Pseudoinverse method for inverse kinematics [22], [23] with the EE trajectory control using Proportional Derivatives (PD) [24], [25].

Previously, Buzz [26] proposed the ClampMag method for EE trajectory control. This method is very simple and fast though the movement of EE is not smooth. Elawady [27] proposed better methods, namely, second order sliding mode-based inverse kinematics (SOSMIK) strategy and continuous second order sliding mode-adaptive inverse kinematics (CSOSM-AIK). SOSMIK can produce a chattering effect, and CSOSM-AIK can produce smooth movements. Whereas in this study, route control uses PD control, which is simpler than CSOSM-AIK, but with the right design, it can produce smooth continuous action.

The contribution of this research is in the strategy of implementing the PD control system for end effector trajectory control based on the inverse kinematics Pseudoinverse Jacobian to get smooth motion and optimal performance for object sorting with digital image input.

This article will be divided as follows. Section II describes the object detection system and the projected coordinates of the detected object on the digital image to its true state. Section III describes the kinematics system used in the manipulator. Section IV presents the mechanical design and framework of the system. Section V provides a report on the results of testing and analysis to validate the performance of the proposed system. And finally, Section VI contains the conclusions from the testing and analysis as a contribution to this research.

## II. MANIPULATOR KINEMATICS

There are two main coordinate system modes in manipulator kinematics modeling, namely cartesian and quartenion. The transformation between two cartesian coordinate systems can be decomposed into rotation and translation.

There are many methods for representing rotation, such as Euler angles, Gibss vectors, and others, but the homogeneous transformation method of real matrix 4x4 orthonormal is the most widely used in robotics [14]. Denavit & Hartenberg [21] proved that the general transformation between two joints requires four parameters, hereinafter known as the Denavit-Hartenberg parameter and this parameter becomes the standard for describing robotic kinematics [14].

### A. Forward Kinematics

This study used the Denavit-Hartenberg (DH) parameter to perform the forward kinematics function. The DH method uses 4 parameters, namely θ, α, d, a. for a robot with i-DOF, the four parameters are determined up to the i-th. The explanation:
1. $\theta i$ is the angle of rotation on the $z_{i-1}$ axis.
2. $\alpha i$ is the angle of rotation on the $x_i$ axis.
3. $d_i$ is the translation on the $z_{i-1}$ axis.
4. $a i$ is the length of the i-th arm.

The DH Matrix is expressed as (1)

$$T_i^{i-1} = \begin{bmatrix} \cos\theta_i & -\cos\alpha_i \sin\theta_i & \sin\alpha_i \sin\theta_i & a_i \cos\theta_i \\ \sin\theta_i & \cos\alpha_i \cos\theta_i & -\sin\alpha_i \cos\theta_i & a_i \sin\theta_i \\ 0 & \sin\alpha_i & \cos\alpha_i & d_i \\ 0 & 0 & 0 & 1 \end{bmatrix} \quad (1)$$

$$T_i^o = \prod_o^i T_i^{i-1}, i > o \quad (2)$$

$i = 1, 2, 3 \ldots; o = 0, 1, 2, 3 \ldots o - 1$
$I$ = joint ith and o = base frame
for example: $(i, o) = (4,0), \therefore T_4^0 = T_1^0 T_2^1 T_3^2 T_4^3$
the final result of the multiplication of a homogeneous matrix in (3)

$$T_n^i = \begin{bmatrix} r_{11} & r_{12} & r_{13} & p_x \\ r_{21} & r_{22} & r_{23} & p_y \\ r_{31} & r_{32} & r_{33} & p_z \\ 0 & 0 & 0 & 1 \end{bmatrix} \quad (3)$$

then the final position of end effector (EE) against base i is $(p_x, p_y, p_z)$.

**B. Inverse Kinematics**

There are two techniques that are usually used to solve inverse kinematics problems in general, namely the analytical method and the numerical method (iterative optimization). Analytical methods are carried out using algebraic and geometric approaches. However, this approach becomes difficult and slow for more complex kinematics problems. Numerically, the iterative method approach can provide a better solution [28].

There are many numerical methods for solving inverse kinematics problems, among which the most popular is Cyclic Coordinate Descent (CCD) [29] and Jacobian invers. CCD is a heuristic method without matrix manipulation, fast computation but prone to redundancy and singularity. CCD is limited to serial kinematics [30]. The solution with the inverse Jacobian is that it linearly models the final effector motion relative to the instantaneous change of the joint angles. Several different ways have been developed to calculate the inverse Jacobian, such as the Jacobian Transpose [31], Pseudo Inverse [22], [23], Damped Least Squares (DLS), Damped Least Squares with Singular Value Decomposition (SVD-DLS), and Selectively Damped Least Squares (SDLS) [26].

In this study, the Jacobian Pseudo Inverse approach is used by considering that this approach can produce an inverse Jacobian with a minimal configuration [32]. The manipulator consists of 4 frames and 4 revolute joints, with the 4th joint designed to have a linear relation j4 = 360 - (j2 + j3). The Jacobian Pseudoinverse method is given an additional system to protect the singularity by limiting the size of the joint angle.

The Jacobian Pseudoinverse is described as follows. If $s$ is the position of EE, is a function of and using the velocity approach, as in (4)

$$\dot{s} = \frac{df(\theta)}{d\theta} \dot{\theta}$$

$$\dot{s} = J\dot{\theta} \quad (4)$$

J is a Jacobian matrix, it is a function, not a constant. So that the inverse kinematics solution as in (5)

$$\dot{\theta} = J^{-1} \dot{s} \quad (5)$$

It is assumed that the expected position is $s_d$, the initial position $s_t$, the process is repeated until the minimum error function is obtained as (6)

$$(s_t - s_d)^2 \approx 0 \quad (6)$$

If $s = [x \ y \ z]^T$ and $\theta = [\theta_1 \ \theta_2 \ \ldots \ \theta_n]^T$ then the Jacobian matrix and linear velocity EE are (7)-(8)

$$J = \begin{bmatrix} \frac{\partial x}{\partial \theta_1} & \frac{\partial x}{\partial \theta_2} & \ldots & \ldots & \frac{\partial x}{\partial \theta_n} \\ \frac{\partial y}{\partial \theta_1} & \frac{\partial y}{\partial \theta_2} & \ldots & \ldots & \frac{\partial y}{\partial \theta_n} \\ \frac{\partial z}{\partial \theta_1} & \frac{\partial z}{\partial \theta_2} & \ldots & \ldots & \frac{\partial z}{\partial \theta_n} \end{bmatrix} \quad (7)$$

$$\begin{bmatrix} \dot{x} \\ \dot{y} \\ \dot{z} \end{bmatrix} = J \begin{bmatrix} \dot{\theta}_1 \\ \dot{\theta}_2 \\ \ldots \\ \dot{\theta}_n \end{bmatrix} \quad (8)$$

If the transformation of the matrix is $T_i^o$ and $J = [J_1 \ J_2 \ \cdots \ J_n]$, then joint the i-th revolution as in (9)

$$J_i = \begin{bmatrix} Z_{i-1} \times (O_n - O_{i-1}) \\ Z_{i-1} \end{bmatrix} \quad (9)$$

And if the i-th joint is translational (prismatic), as in (10)

$$J_i = \begin{bmatrix} Z_{i-1} \\ 0 \end{bmatrix} \quad (10)$$

$Z_i$ is the first three elements in the 3rd column of the matrix $T_i^o$, and $O_i$ is the first three elements of the 4th column of the matrix $T_i^o$ [6]. The inverse kinematics solution can be carried out by the approach as (11)

$$\begin{bmatrix} \dot{\theta}_1 \\ \dot{\theta}_2 \\ \ldots \\ \dot{\theta}_n \end{bmatrix} = J^{-1} \begin{bmatrix} \dot{x} \\ \dot{y} \\ \dot{z} \end{bmatrix} \quad (11)$$

With the Pseudoinverse approach, as in (12)

$$\begin{bmatrix} \dot{\theta}_1 \\ \dot{\theta}_2 \\ \ldots \\ \dot{\theta}_n \end{bmatrix} = J^+ \begin{bmatrix} \dot{x} \\ \dot{y} \\ \dot{z} \end{bmatrix} \quad (12)$$

The Jacobian matrix pseudoinverse approach or also known as the Moore-Penrose inverse [33], [34] by computing Singular Value Decomposition (SVD) as in (13).

$$J = USV$$
$$J^+ = V_1 S_1^{-1} U_1^T \quad (13)$$

Substitute $J^+$ (13) to (12).

$$\begin{bmatrix} \dot{\theta}_1 \\ \dot{\theta}_2 \\ \ldots \\ \dot{\theta}_n \end{bmatrix} = V_1 S_1^{-1} U_1^T \begin{bmatrix} \dot{x} \\ \dot{y} \\ \dot{z} \end{bmatrix} \quad (14)$$

## III. OBJECT DETECTION AND COORDINATE PROJECTION

### A. Object Detection

The object detection meant in this research is the detection of the object's color and position. The object's existence was identified using the HSV color segmentation approach [19], [20]. Color perception with HSV is claimed to be more in tune with human color perception compared to other color systems, and HSV is claimed to be more robust against illumination variations [35], [36].

The RGB image from the camera was converted to HSV. Each HSV element was separated and the H element was multiplied by 360. As long as S and V are in the 0.1-1 range, pixels are defined as red pixels if H ≤ 20 or ≥ 280. Then, pixels are classified as Blue if H is in the range 90-120, and Green if H is in the range 130-160. For each pixel that had been classified further, each pixel according to its color was filtered median by neighboring pixels, 3x3 size, so that at least 100 pixels were connected (contour). Connections less than 100 pixels were deleted, for example, the input image in Figure 1(a). The results are shown in Figures 1(b), 1(c), 1(d). Lastly, the area and center of each contour were calculated and the final detection results are shown in Figure 1(e).

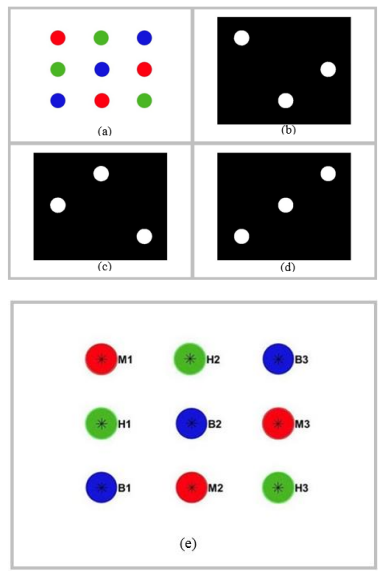

Figure 1. Objects Detection (a)Objects to be Detected, (b) Red Object Detection, (c) Green Object Detection, (d) Blue Object Detection, (e) Final Object Detection Results based on HSV Color

### B. Digital Image Coordinate Projection - Actual

The camera is set at the camera resolution ($w_m$ x $h_m$) 640 x 480 px$^2$. The camera capture center point is parallel to the center point of the object area with a height of about 52 cm from the object area so that the area of the object captured by the camera ($w_o$ x $h_o$) is 50 x 37, 5 cm$^2$.

Digital image coordinate projection equation – actual ($x_o, y_o$)

$$x_o = \frac{w_o}{w_m} \times x_m$$

$$y_o = \frac{h_o}{h_m} \times y_m$$

($x_m, y_m$) = coordinates on a digital image
($x_o, y_o$) = actual coordinates

## IV. MECHANICAL DESIGN AND FRAMEWORK SYSTEM

The used manipulator principle in Figure 2 is equivalent to the 5-DOF model in Figure 3 ($J_n$ is the n-th joint). With the aim of reducing the computational cost of control, using the MeArm model [18] in Figures 2 and 4, the 5-DOF control system is simplified to 3-DOF. The 4[th] joint (j4) without servo is automatically controlled mechanically to always be parallel to the XY surface. The 5[th] joint (servo 4 in Figure 2) as a drive for the gripper position is controlled separately to avoid the possibility of $s$ colliding with objects on the XY surface when moving towards the target object. Servo 5 is used for the function of opening and holding the gripper.

The angle of j4, $\theta_4$

$$\theta_4 = -(\theta_2 + \theta_3)$$

The gripper arm is moved by controlling the angle $\theta_5$, the gripper arm is controlled after s reaches the target, with the situation of the arm $a_4$ is always set parallel to the XY surface,

$$a_{5XY} = a_5 \cos \theta_5$$
$$a_{5Z} = a_5 \sin \theta_5$$
$$a_{4XY} = a_4 + a_{5XY}$$

$a_{5XY}$ = the horizontal distance (XY) of the joint to the gripper grip point.
$a_{5Z}$ = the vertical distance (Z) of the joint to the point of gripper grip.
$a_{4XY}$ = the horizontal distance (XY) of the joint to the gripper grip point j$_4$

$a_{4XY}$ is the value used as a variable value $a$ for n = 4 in the DH parameter. With $a_4$ length = 9 cm, and $a_5 = 10$ cm with the initial angle, then the variable $\theta_5 = 60°$, then variable $a$ for n = 4 on the DH parameter is 14 cm. Table 1 is the DH parameter of the manipulator used, the variable value $\alpha$, $a$, and $d$ are constant.

12 • Indra Agustian, et. al.

TABLE 1
DH PARAMETER

|   |   | Parameter |   |   |   |
|---|---|---|---|---|---|
|   |   | $\Theta$ | $\alpha$ | $a$ | $d$ |
| n | 1 | $\theta 1$ | 90 | 3 | 17.5 |
|   | 2 | $\theta 2$ | 0 | 22.3 | 0 |
|   | 3 | $\theta 3$ | 0 | 31.5 | 0 |
|   | 4 | $\theta 4 = -(\theta 2 + \theta 3)$ | 0 | 14 | 0 |

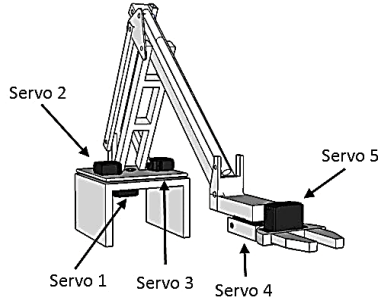

Figure 2. Manipulator Model Design

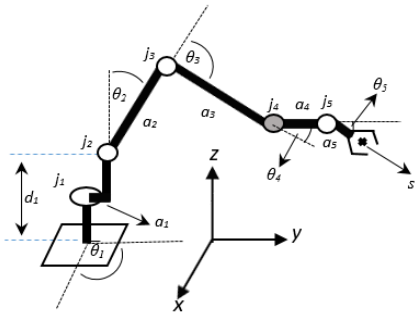

Figure 3. 5 DOF Equivalent Manipulator Model

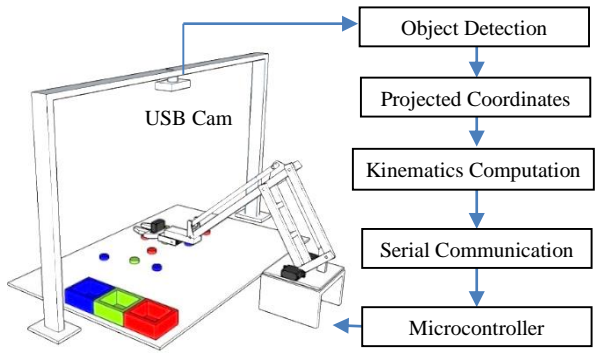

Figure 4. An Experimental Model of Object Sorting Manipulator based on Color

The target of X and Y position, $s_{XY}$
$s_{XY} = Coordinate\ XY\ actual\ detected\ object$
The scenario, after s reaches the target position, $\theta_5$ initial is set to 30°, then it is moved towards -60°. The position of the gripper when gripping the object is 1.5 cm from the surface, then the target position setting s on the Z-axis in cm.

$$s_z = 1.5 + a_{5Z}$$
$$s_z = 1.5 + |a_5 \sin\theta_5|$$
$$s_z = 1.5 + |13 \sin(-60)|$$
$$s_z = 12.76$$

The system scenario is shown in Figure 4. Based on color data and object detection location, the robot arm manipulator is controlled to perform the object sorting process, pick up and place objects based on color. The manipulator moves from the initial state towards the red object sequentially, returning to the initial position upon completion. To avoid changing the position data every after moving an object, the object detection process is repeated for the next object moving process, the process is declared complete after no more objects are detected. The joint angle in degrees $(\theta_1, \theta_2, \theta_3, \theta_4, \theta_5)$ for the initial positions (x, y, z) in cm is (120, 93, -132, 39, 60). The initial cartesian position and the object placement position are shown in Table 2.

Singularity is avoided by giving a maximum limit to the target end effector to avoid the possibility of a link being parallel to the previous link, by giving a minimum - maximum angle between joints to avoid redundancy if there is a large angle at the maximum or minimum $\Delta\theta$ given at the angle is 0 and continues with the next iteration. Table 3 and Table 4 are the limits given to avoid the occurrence of a singularity.

The manipulator trajectory control scenario is shown in the diagram in Figure 5, where the EE position is controlled using the Pseudoinverse Jacobian approach. In the implementation, $\dot{s} = J\dot{\theta}$ is discretised to $\Delta s = J\Delta\theta$. $\Delta s$ can be assumed as linear velocity EE linear and $\Delta\theta$ assumed as angular velocity. The linear velocity of EE is controlled with a proportional derivative (PD) control with (15)-(17).

TABLE 2
INITIAL POSITIONS AND PLACEMENT OF OBJECTS

| Position | x | y | z |
|---|---|---|---|
| Initial | -20 | 35 | 20 |
| Red | -20 | 35 | 20 |
| Green | -20 | 45 | 20 |
| Blue | -20 | 55 | 20 |

TABLE 3
END EFFECTOR POSITION RADIUS LIMITATION

| Position | Minimum | Maximum |
|---|---|---|
| $x$ | -40 cm | 40 cm |
| $y$ | 20 cm | 60 cm |
| $z$ | 10 cm | 60 cm |

TABLE 4
JOIN ANGLE CONSTRAINTS

| Join | Minimum | Maximum |
|---|---|---|
| $\theta_1$ | 20° | 160° |
| $\theta_2$ | 30° | -80° |
| $\theta_3$ | -20° | -80° |

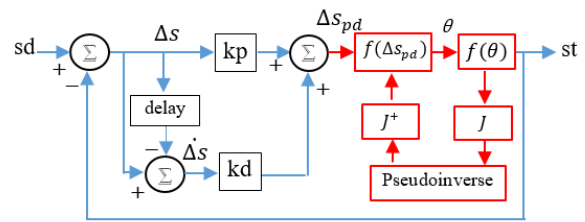

Figure 5. PD Manipulator Control Diagram

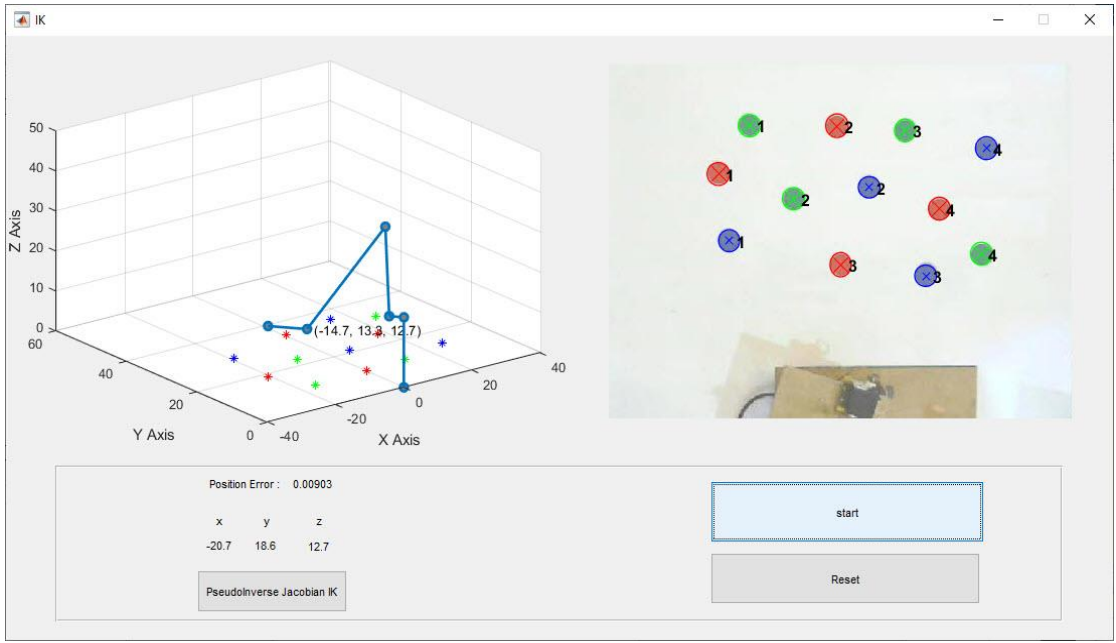

Figure 6. Manipulator Control Panel

$$\Delta s_{PD} = k_p \cdot \Delta s + k_d \cdot \dot{\Delta s} \cdot t \quad (15)$$

$$\Delta \theta = J^+ \Delta s_{PD}$$

$$\Delta \theta = J^+ (k_p \Delta s + k_d \dot{\Delta s}) \quad (16)$$

$$\theta_t = \theta_{t-1} + \Delta \theta$$

$$\theta_t = \theta_{t-1} + J^+ (\Delta s + k_d \dot{\Delta s}) \quad (17)$$

$$s \in \mathbb{R}^3, J \in \mathbb{R}^{4 \times 4}, \theta \in \mathbb{R}^4$$

The hardware and software components utilized in this study are: RDS3135 Coreless Servo Motor with 35 kg.cm of torque used on each joint manipulator, MG996R Micro Servo for gripper on end effectors, Arduino Uno microcontroller as servo control interface, Logitech C310 USB Camera and object detection, and kinematic computation programs using the Matlab R2018a program run on an Intel i5-2410M laptop CPU @ 2.3GHz 8GB RAM with Windows 64-bit Operating System.

## V. RESULTS AND DISCUSSION

The results of the control panel GUI design on the PC are shown in Figure 6. The GUI displays the results of object detection in the image, visual simulation of the robot arm, EE coordinates, EE position error (scalar) against the target. The results of the robot arm manipulator design are shown in Figure 7. First, the start button was executed, the system would connect to the microcontroller as the interface for controlling the joint motion of the robot arm, after a successful connection, the camera started recording images of target objects and the PC performed the object detection process, color and position as in Figure 8. Object detection was tested at a lighting level of 500-1000 lux. The color and position data are displayed on the robot arm simulation graph, then the system performed an inverse kinematics simulation. The angular data of each joint of the simulation results was sent to the microcontroller interface that drives the robot arm and run the object sorting process based on color automatically.

In the test, 12 objects (4 red, 4 green, 4 blue) were randomly placed on the robot arm work table. Table 5 shows the color detection result data, the position of the object in the image, and the mapping of the object's actual location. M1, M2, M3, M4 are red objects; H1, H2, H3, H4, are green objects; B1, B2, B3, B4 are blue objects.

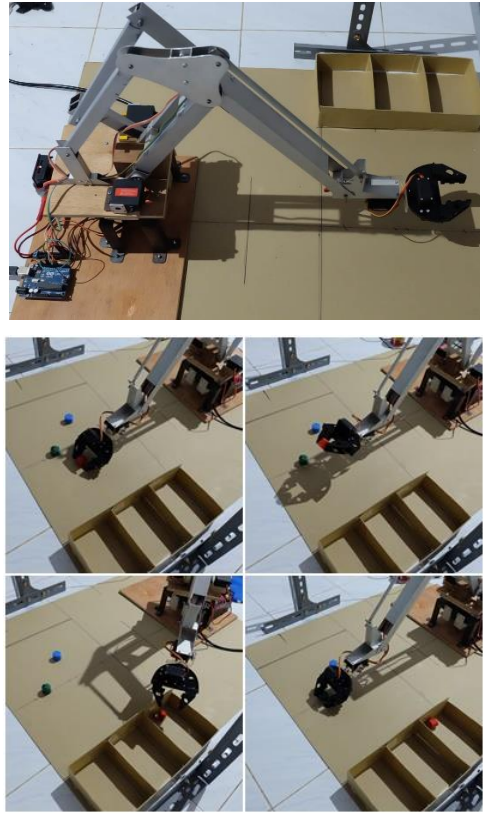

Figure 7. Hardware Manipulator and Sample Experiment

TABLE 5
THE RESULT OF COLOR DETECTION AND OBJECT POSITION

| No. | Object | Position on digital image (px) | | Mapping to actual position (cm) | |
|---|---|---|---|---|---|
| | | x | y | x | y |
| 1 | M1 | 152.20 | 148.73 | -10.99 | 49.70 |
| 2 | M2 | 316.04 | 84.49 | -0.75 | 53.72 |
| 3 | M3 | 321.66 | 271.90 | -0.40 | 42.01 |
| 4 | M4 | 457.26 | 196.63 | 8.08 | 46.71 |
| 5 | H1 | 194.99 | 83.06 | -8.31 | 53.81 |
| 6 | H2 | 254.80 | 182.75 | -4.57 | 47.58 |
| 7 | H3 | 410.32 | 89.99 | 5.15 | 53.38 |
| 8 | H4 | 515.76 | 257.85 | 11.74 | 42.88 |
| 9 | B1 | 166.68 | 239.52 | -10.08 | 44.03 |
| 10 | B2 | 360.53 | 166.94 | 2.03 | 48.57 |
| 11 | B3 | 438.88 | 286.18 | 6.93 | 41.11 |
| 12 | B4 | 522.39 | 114.16 | 12.15 | 51.86 |

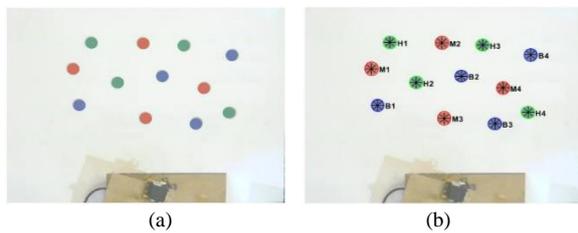

Figure 8. (a) The Digital Image of The Object Captured by The Camera, (b) The Object Detection Result

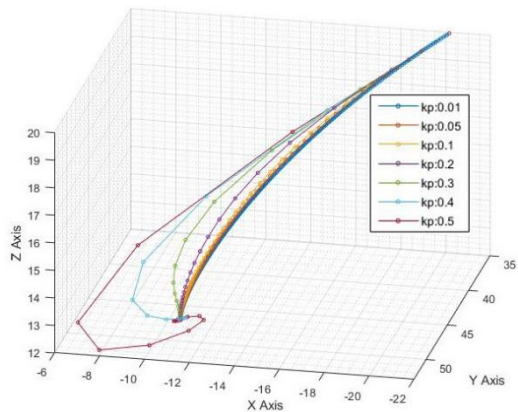

Figure 9. EE movement with the variation of $k_p$

With proportional constants $(k_p) = 0.1$ and derivative constants $(k_d) = 0.01$. Constants were selected by trial and error approach. $k_p \cdot \Delta s$ basically, partitioned the error, $k_p = 0.1$ meaning that the error was partitioned into 10, the smaller the value of the $k_p$ path, the closer it was to a straight line, but this would enlarge the computation considering that the total iteration would increase significantly. The greater the $k_p$, the less the number of iterations. Thus, even though the computation is faster, the motion is coarser and might result in overshoot as seen in some samples of the Kp tuning test in Figure 9 and Figure 10. $k_p$ 0.01 has the smoothest movement, yet the slowest converge. Meanwhile, $k_p$ 0.5 has the overshoot and indented movement. Based on this data, $k_p$ 0.1 is considered ideal for this study. As seen in Figure 11, where all of the kd values are considered adequate, this study uses $k_p = 0.1$ and $k_d = 0.01$. The constant is applicable for all the objects on the workspace.

The test results are shown in Figure 12. The EE trajectory of the robot arm from the initial position to the location of the first red target object (RT) was picked then taken to the location (x, y, z) ≈ (-20, 35, 20). The gripper released the object towards the red object placement box. Figure M1 shows the iteration response to the error position distance EE to the target at M1.

The node in Figure 13 shows the position of EE in the n-th iteration, the smaller the position error of EE towards the target, the distance change between the iterations is getting smaller, meaning that there is a decrease in speed until it finally stops and takes or releases the object.

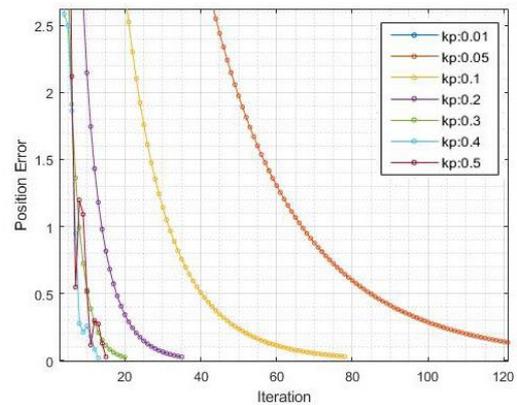

Figure 10. EE Position Error vs Number of Iterations with The Variation of $k_p$

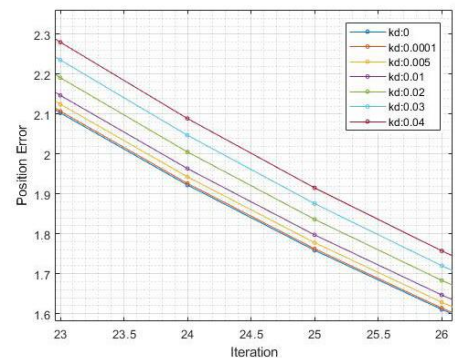

Figure 11. EE Position Error vs Number of Iterations at $k_p = 0.1$ with The Variation of $k_d$

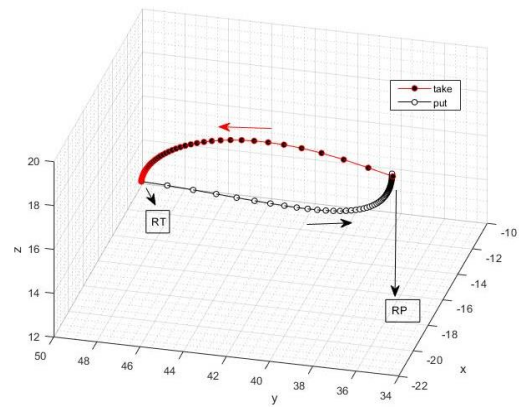

Figure 12. EE Trajectory for Action on Object M1

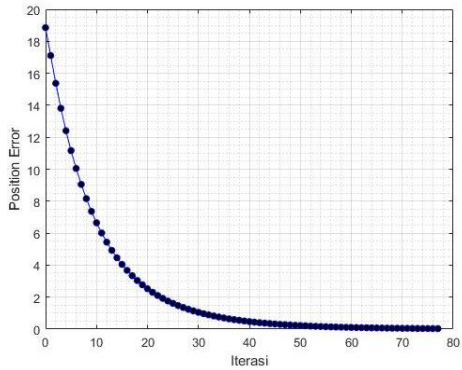

Figure 13. EE Position Error vs Number of Iterations for Actions Corresponding to Object M1

The example of EE trajectory in the first pick-place task for red, green, and blue objects is shown in Figure 14(a)(b)(c). No abnormal EE movement is seen. This indicates that the inverse kinematics system applied does not cause a singularity. From the total iteration data, the initial error and the final error for each task in the first test are shown in Table 6. It is found that with an average initial error of 24.99 cm, an average final error of 0.0305 cm is obtained, the average requires 82 iterations. The largest number of iterations is 93 with an initial error of 22.05 cm and 74 iterations for an initial error of 27.82 cm. This indicates that the number of iterations is not only determined by the initial error size.

From Figure 15(a), it can be seen that the iteration distribution to the error position is less than 10 cm. For a distance of ≈ 1 cm, a minimum of 43 iterations and a maximum of 58 iterations are needed. And from Figure 15(b), for a distance of less than 0.1 cm at least 16 iterations are necessary.

TABLE 6
TOTAL ITERATION DATA AND POSITION ERROR FROM THE SIMULATION

| No. | Object | Step | Total | Initial | Final |
|---|---|---|---|---|---|
| 1 | M1 | 0 | 78 | 18.85 | 0.0305 |
| 2 |  | 1 | 92 | 18.72 | 0.0304 |
| 3 | M2 | 0 | 74 | 27.82 | 0.0305 |
| 4 |  | 1 | 91 | 27.82 | 0.0315 |
| 5 | M3 | 0 | 93 | 22.05 | 0.0300 |
| 6 |  | 1 | 91 | 22.05 | 0.0302 |
| 7 | M4 | 0 | 84 | 31.28 | 0.0306 |
| 8 |  | 1 | 91 | 31.28 | 0.0305 |
| 9 | H1 | 0 | 73 | 23.31 | 0.0301 |
| 10 |  | 1 | 77 | 16.34 | 0.0300 |
| 11 | H2 | 0 | 81 | 17.25 | 0.0299 |
| 12 |  | 1 | 78 | 17.25 | 0.0313 |
| 13 | H3 | 0 | 73 | 27.48 | 0.0308 |
| 14 |  | 1 | 78 | 27.48 | 0.0297 |
| 15 | H4 | 0 | 86 | 32.62 | 0.0309 |
| 16 |  | 1 | 79 | 32.62 | 0.0312 |
| 17 | B1 | 0 | 84 | 12.33 | 0.0307 |
| 18 |  | 1 | 76 | 16.48 | 0.0310 |
| 19 | B2 | 0 | 81 | 24.08 | 0.0302 |
| 20 |  | 1 | 76 | 24.08 | 0.0307 |
| 21 | B3 | 0 | 90 | 31.16 | 0.0301 |
| 22 |  | 1 | 78 | 31.16 | 0.0306 |
| 23 | B4 | 0 | 75 | 33.11 | 0.0313 |
| 24 |  | 1 | 76 | 33.11 | 0.0302 |
|  | Mean |  | 82 | 24.99 | 0.0305 |

Step:
    0 = picking motion
    1 = placing motion

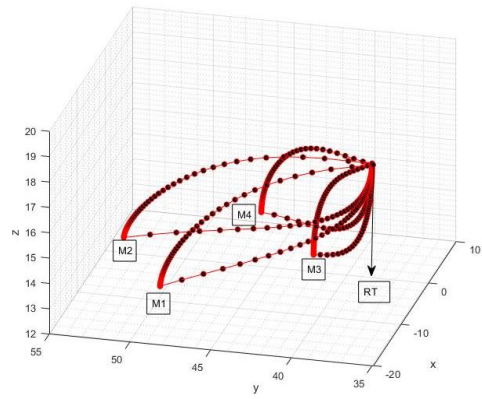

(a)

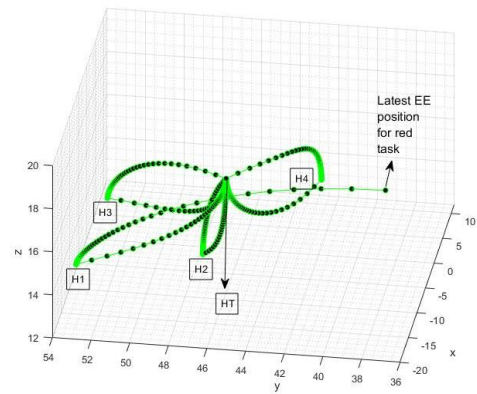

(b)

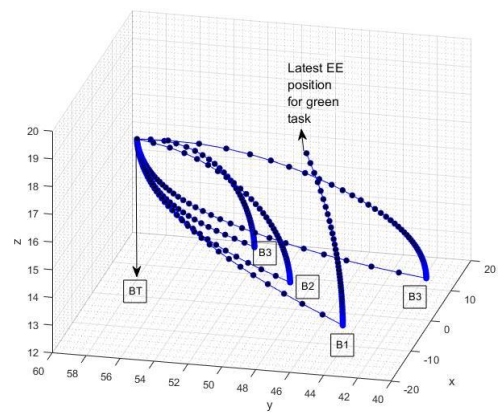

(c)

Figure 14. The Sample EE Trajectory on The Sorting Action of (a) Red, (b) Blue, (c) Blue Objects

The test results in Table 7 show that the EE movement is still within the limits required in Table 3, the radius of the EE position. From Figure 16, it is indicated that the angle of each joint is still within the limits required in Table 4. These two data indicate that the manipulator does not experience a singularity.

TABLE 7
TRAJECTORY OF MINIMUM MAXIMUM END EFFECTOR POSITION

| xmin | xmax | ymin | ymax | zmin | zmax |
|---|---|---|---|---|---|
| -20.16 | 12.15 | 34.91 | 55.00 | 19.97 | 12.73 |

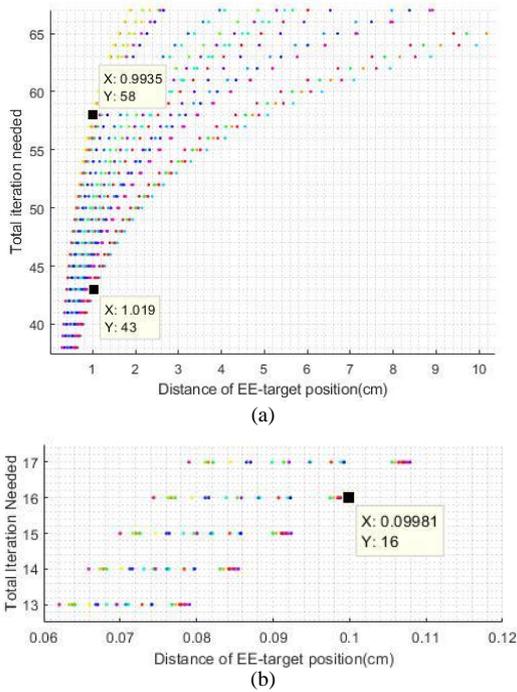

(a)

(b)

Figure 15. Iteration vs Error Distribution (a) Original Scale (b) Zoom Version of (a)

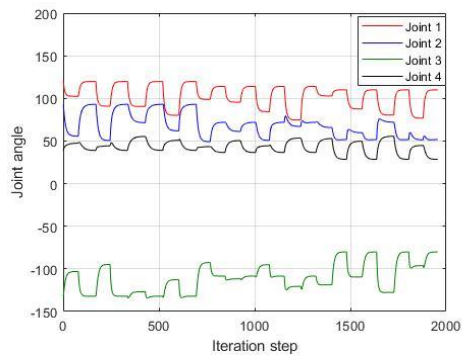

Figure 16. Route of Joint Angle to Iteration

TABLE 8
FINAL ERROR STATISTICS OF THE TEST

| EE-target's | Mean | Standar | Max | Min |
|---|---|---|---|---|
| x | -0.0086 | 0.41259 | 1.40885 | -1.596 |
| y | -0.0047 | 0.22245 | 0.76329 | -0.8509 |
| z | -0.0022 | 0.52336 | 2.00255 | -1.9053 |
| Euclidean xy | 0.39882 | 0.24635 | 1.60043 | 0.01067 |
| Euclidean xyz | 0.62397 | 0.32284 | 2.15026 | 0.02186 |

TABLE 9
FIRST ATTEMPT PICK AND PLACE OBJECT RATE

| Object | Pick | | Put | |
|---|---|---|---|---|
| | Success | Failure | Success | Failure |
| Red | 96.00% | 4.00% | 100.00% | 0.00% |
| Green | 94.00% | 6.00% | 100.00% | 0.00% |
| Blue | 92.75% | 7.25% | 100.00% | 0.00% |
| Total | 94.25% | 5.75% | 100.00% | 0.00% |

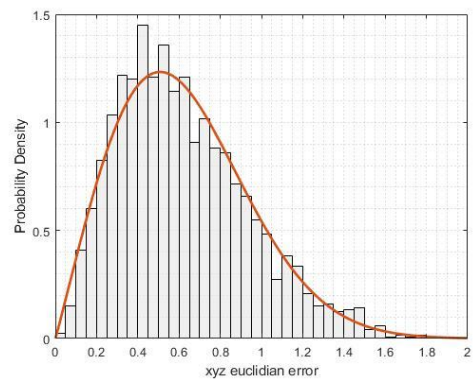

Figure 17. PDF xyz Final Error of Actual Manipulator Motion, bin = 0.05

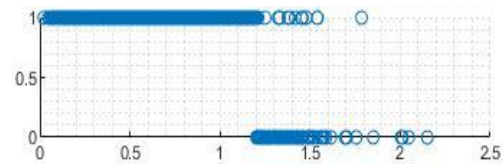

Figure 18. Graph of Object Fetching Success.

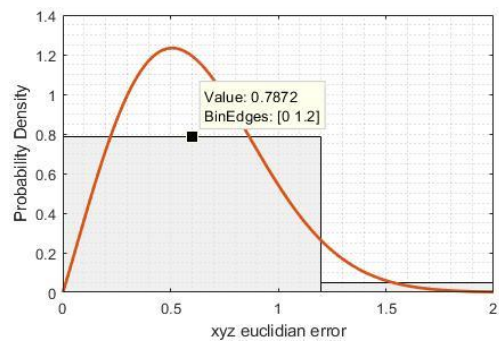

Figure 19. PDF of The Euclidean xyz Distance Error

For the implementation of pick-place objects task by the actual manipulator, 100 trials were carried out, each trial consisting of 12 tubular objects with 2 cm in diameter and 1 cm in height, divided by four red objects, four green objects, and four blue objects. So, there were a total of 1,200 picks and 1,200 places of objects, totaling 2400 of routes. Since the dimensions of the tip of the gripper is 0.5 cm, the distance between objects must be above 0.5 cm. In order for this parameter to not interfere with the success of capturing the object, the distance between objects during testing was set to be at least 1 cm.

The final statistical data of actual manipulator implementation error on simulation data is shown in Table 8. From a total of 2400 routes, the average euclidean error xyz = 0.6240, standard deviation = 0.3228 with a maximum xyz error of 2.15026 cm and a minimum of 0.02186 cm. Figure 17 shows a histogram with a normalization probability density function (PDF) and a Weibull probability distribution [37] with a scale parameter of 0.7059, parameter of shape 2.0450, and forming a right-skewed distribution. Parameter shape 2.0450> 1.0 indicates that the success rate is still quite effective.

The experimental results in Table 9 show the experimental statistical data without correction of the success of the process. The objects that failed to be taken were separated, and the percentage of the success rate of taking the object reached 94.25%. Figure 18 shows a graph of the results of the successful experiment based on the Euclidean xyz EE - target distance, the vertical axis 0 = failure, 1 = success. The robot is 100% successful in picking up objects at the Euclidean xyz EE-target distance of less than 1.2 cm, above 1.2 cm the robot failed to pick up objects amounting to 51.88% of the total 133 attempts.

Based on the PDF data xyz final error from Figure 17, if the big histogram bin width is equal to 1.2 cm, that is the error distance with 100% pick success, Figure 19 is obtained, then the probability of the success rate if the xyz error <1.2 cm is $0.7872 \times (1.2 - 0)$, which is equal to 0.9446 or 94.46%.

The average process of detecting 12 objects at once at a camera resolution of 640 x 480 px$^2$ was 0.263 s and for one motion pick or place takes an average of 6 s with 0.072 s average time consumption for one kinematics iteration. To shorten the pick-place time can be done by adjusting $k_p$ and $k_d$ to increasing the speed. However, due to the focus of this study on manipulator kinematics, speed optimization was not studied in detail.

## VI.  CONCLUSIONS

This study developed a hardware manipulator model and implementation for sorting purposes. The manipulator navigation used forward kinematics DH and inverse kinematics Pseudoinverse Jacobian with Proportional-Derivative (PD) control. From the testing of the developed model and hardware, the PDF data and the Weibull probability distribution show that the proposed manipulator navigation system has good performance and the manipulator does not experience any singularity at all. The success rate of the sorting implementation reached 94.46% with an error of the euclidean distance less than 1.2 cm. In the future, this kinematics navigation system can be used for different implementations. Another further research that can be done is kinematics to tracking dynamic objects with different attributes.

## ACKNOWLEDGMENT

This research is the first part of the Smart-robot weed control research supported by the University of Bengkulu through a development research program with a contract No. 2050 / UN30.15 / PG / 2020.